\documentclass[11pt,a4paper]{article}
\usepackage{SyntaxFest2019}
\usepackage{times}
\usepackage{latexsym}
\usepackage{amsmath}
\usepackage{graphicx}
\usepackage{url}
\usepackage{subfig}
\usepackage{float}
\usepackage{adjustbox}
\usepackage{xcolor}
\usepackage{mathptmx}
\usepackage{expex}
\usepackage[title]{appendix}
\usepackage{algorithm,amsmath,algpseudocode}

\usepackage{enumitem}
\setlist[itemize]{noitemsep, label={\large\textbullet}}

\title{Artificially Evolved Chunks for Morphosyntactic Analysis}

\author{Mark Anderson \qquad  David Vilares \qquad Carlos G{\'o}mez-Rodr{\'i}guez \\
  Universidade da Coru\~na, CITIC \\
  FASTPARSE Lab, LyS Research Group, Departamento de Computaci\'on \\
  Campus Elvi\~{n}a, s/n, 15071\\ 
  A Coru\~{n}a, Spain\\
  {\tt \{m.anderson,david.vilares,carlos.gomez\}@udc.es}}
  
\date{}
\newcommand{\david}[1]{\textcolor{black}{#1}}
\newcommand{\markda}[1]{\textcolor{black}{#1}}

\newcommand{\carlos}[1]{\textcolor{black}{#1}}
\begin{document}
\maketitle
\begin{abstract}
We introduce a \markda{language-agnostic} evolutionary technique for automatically extracting chunks from dependency treebanks. We evaluate these chunks on a number of morphosyntactic tasks, namely POS\footnote{POS tagging is used throughout to refer to universal part-of-speech (UPOS) tagging.} tagging, morphological feature tagging, and dependency parsing. We test the utility of these chunks in a host of different ways. We first \david{learn} chunking as one task in a shared multi-task framework \david{together} with POS and morphological \markda{feature} tagging. The predictions from this 
network are then used as input to augment sequence-labelling \markda{dependency} parsing. Finally, we investigate the impact chunks have on 
\markda{dependency} parsing in a multi-task framework. Our results from these analyses show that \markda{these} chunks improve performance at different levels of syntactic abstraction on English UD treebanks and a small, diverse subset of non-English \markda{UD} treebanks.
\end{abstract}
\section{Introduction}
Shallow parsing, or chunking, consists of identifying constituent phrases \cite{Abney1997}. As such, it is fundamentally associated with constituency parsing\carlos{, as it can be used as a first step for finding a full constituency tree \cite{ciravegna-lavelli-1999-full,tsuruoka-tsujii-2005-chunk}.} 
\carlos{However, chunking information can also be beneficial for dependency parsing \cite{AttardiDellOrletta2008,tammewar2015can}, and vice versa \cite{kutlu2016noun}.}
Latterly, \newcite{lacroix2018investigating} explored the efficacy of noun phrase (NP) chunking with respect to universal dependency (UD) parsing and POS tagging for English treebanks. As UD treebanks do not contain chunking annotation, they deduced chunks by adopting linguistic-based phrase rules. 
They observed improvements on POS and morphological feature tagging \markda{in a} 
shared multi-task framework for the 
English treebanks in UD version 2.1 \cite{ud21}. 
\markda{However, an increase in performance for } 
\markda{parsing was only obtained for one treebank.}


\paragraph{Contribution} 1. We first relax the standard definition of chunks and present an evolutionary method to automatically deduce chunks for any language given a dependency treebank. 2. We show that chunking information can improve performances for POS tagging, morphological feature tagging, and dependency parsing, both in a multi-task and a single-task framework.

\section{Chunks \carlos{and chunking rules}}

\begin{figure}[htbp!]
    \centering
    \includegraphics[width=0.435\linewidth]{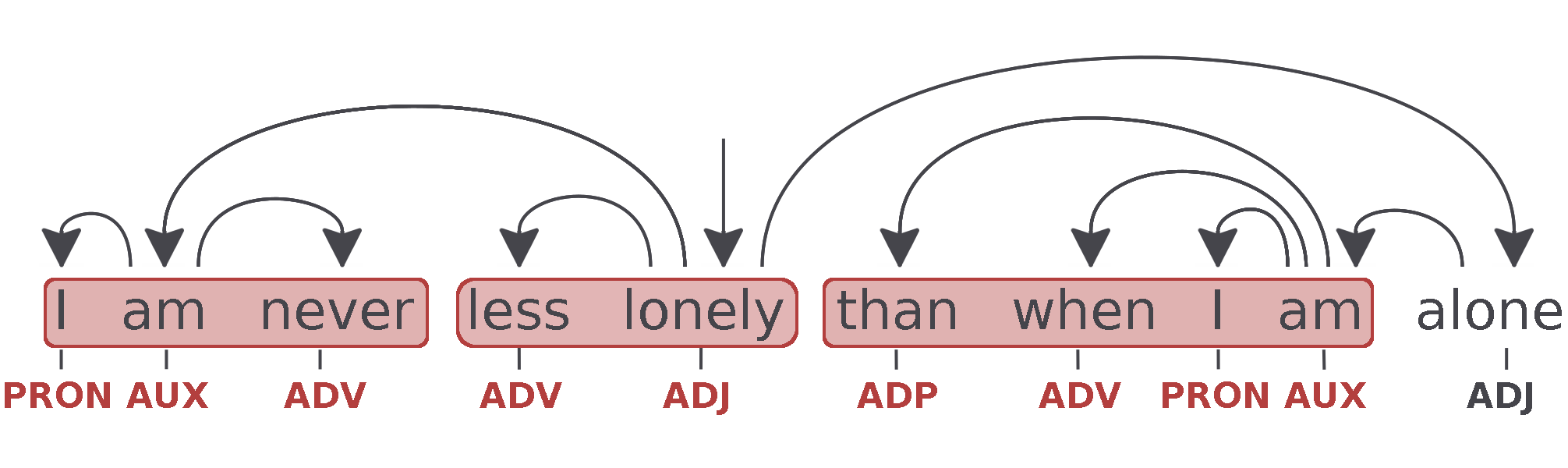}
    \caption{Candidate phrase rules are extracted 
    by selecting subtrees with one level of dependency.}
    \label{fig:chunk_extract}
\end{figure}

\carlos{While \newcite{lacroix2018investigating} described a method to obtain chunks from sentences with UD annotations, their approach is limited to NP chunks and requires hand-crafted linguistic rules, meaning that it cannot be transferred to other languages without language-specific knowledge. In contrast, we introduce a fully automatic approach to obtain chunks from UD-annotated sentences in a 
language-agnostic way}.
Figure \ref{fig:chunk_extract} depicts our method of extracting candidate chunk types.
\paragraph{Chunk definition} Here we loosen the definition of a chunk and consider any base-level subtree a possible chunk \markda{defined by the following criteria: \david{(i)} the components of a chunk are syntactically linked; \david{(ii)} there is only one level of dependency (one head and its dependents); \david{(iii)} the components are continuous; and \david{(iv)} no \markda{dependents} within a chunk has a dependent outside the chunk.}



\paragraph{Describing chunks with rules} For each subtree \carlos{in the training set} that meets the above criteria, the corresponding \carlos{sequence of} POS tags of its words is saved as a candidate rule. Each rule is collected for a given treebank to construct a ruleset of unique candidate chunk types. \markda{When more than one \carlos{overlapping} subtree meets these conditions the maximal substring is used, e.g. in Figure \ref{fig:chunk_extract} PRON AUX ADV is chosen instead of PRON AUX or AUX ADV. }  \markda{We allow any chunk type with the exception of those containing the PUNC POS tag and we apply a mild frequency cut of 5 to make the problem more tractable.} The English-EWT treebank, for example, results in a ruleset consisting of 512 candidates.

\paragraph{Annotating with rulesets} \carlos{This ruleset (or any subset of it) can be applied to a UD \markda{treebank} 
to obtain chunks, by using them as patterns that generate a chunk when they are matched by a sequence of POS tags \markda{and meet the criteria described above}.\footnote{Rules are applied from longer (more specific) to shorter (more generic).} In particular, we can apply it to the training set to obtain a set of chunks on which to train a statistical chunker to process arbitrary texts and help morphosyntactic tasks.} \markda{When annotating a treebank, the POS tag of the head is used as a suffix for the chunk type, e.g. DET ADJ NOUN would result in IOB tags of B-NOUN and I-NOUN, assuming the head of this phrase corresponds to the NOUN tag \cite{ramshaw1999text}}.


However, not all candidate rules are useful and can impact the ability of a chunker to make sensible predictions.
\carlos{For this reason, we will not use the whole candidate ruleset obtained from a training corpus, but instead try to find a subset of the ruleset whose resulting set of chunks strikes a good balance between the following criteria: (i) coverage (i.e. there should be enough chunks to maximize their informativeness for morphosyntactic tasks) and (ii) consistency and learnability (i.e. the chunks should follow patterns predictable enough to be easily learnable by a machine learning model, so that our approach is not undermined by low chunking accuracy). Our hypothesis is that these two characteristics (which we quantify with a fitness function in the next section) are reasonable proxies for the usefulness of a particular set of chunks for morphosyntactic tasks.}

\carlos{Note that 
to achieve this,} it is not possible to merely remove error-prone rules from the ruleset because there is a complicated interplay between rules, 
i.e. if the 10\% most error-prone rules are removed, the overall accuracy of the system is not guaranteed to improve. Furthermore, with so many candidate rules, it is not possible to try every combination as this results in an astronomical number ($2^n$). Therefore, we aim to use an evolutionary method to find optimal subsets of rules to be used when annotating treebanks. 

\section{Evolutionary search for chunk rules} 
Evolutionary algorithms aim to optimise an objective (fitness) function by evaluating a population of individuals and subsequently generating a new population based on the best performing individuals from the population \cite{back1996evolutionary}. This process is then repeated until a set number of generations is reached or until the fitness function converges. Each individual consists of a set of parameters and its corresponding objective function value, or fitness. The fitness of an individual is used to decide whether to use it as a parent for subsequent generations or to remove it from the population. We introduce 
the techniques used to select parents and how they are \markda{then} used to generate \markda{offspring} \carlos{(Algorithm \ref{evo_algo} in Appendix \ref{appendix:algo})}.
\paragraph{K-best parent selection}The selection operator makes the population converge. 
We used the simple k-best method where the top k individuals of a population are selected as the parents.

\paragraph{Mutation} Mutation is a genetic operator which prevents a population becoming too genetically similar by randomly altering individuals. This ensures that at least some level of genetic diversity is maintained from generation to generation. Our individuals have binary genes, so our mutation operator flips each gene with a probability P$_{\textrm{mutate gene}}$.
\paragraph{Crossover} Crossover is a genetic operator which also preserves genetic variety in a population. In single-point crossover, a random index $\kappa$ is chosen and the substring 0-$\kappa$ of parent$_x$ is replaced with the corresponding part of parent$_y$ and vice-versa. This results in two offspring. Single-point crossover can be extended to x-point crossover, where x points are used to cut individuals.

We used the DEAP framework for our implementation \cite{DEAP_JMLR2012}\carlos{, and the parameters in Table \ref{tab:evo_parameters} (Appendix \ref{appendix:hyper})}. 
We represented our rulesets as a binary vector, where 1 meant a rule was used and 0 meant it was not. Our fitness function was obtained by combining the F1-score of a chunker implemented with the sequence-labelling framework NCRF++ \cite{yang2018ncrf} and the proportion of the maximum compression rate, weighted 1.0 and 0.5 respectively. 
The compression rate, $r$, is defined as:
\begin{equation}
    r = \frac{C_{\textrm{tokens}}}{C_{\textrm{chunks}}+C_{\textrm{out}}}
\end{equation}
where $C_{\textrm{tokens}}$ is the number of tokens in a treebank, $C_{\textrm{chunks}}$ the number of chunks a ruleset creates, and $C_{\textrm{out}}$ the number of tokens outside of chunks. And subsequently the proportion of the maximum compression rate, $r\%$ is defined as:
\begin{equation}
    r\% = \frac{r_{\textrm{subset}}-1}{r_{\textrm{all}}-1}
\end{equation}
where $r_{\textrm{subset}}$ is the compression rate of the current rule subset and $r_{\textrm{all}}$ is the compression rate of the full ruleset.

\begin{figure}[htbp!]
    \centering
    \subfloat[English-EWT]{\includegraphics[width=0.4\linewidth]{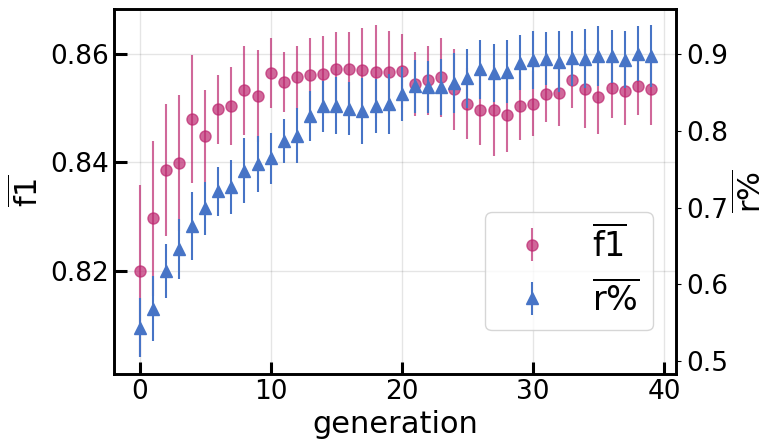}}
    ~
    \subfloat[Japanese-GSD]{\includegraphics[width=0.4\linewidth]{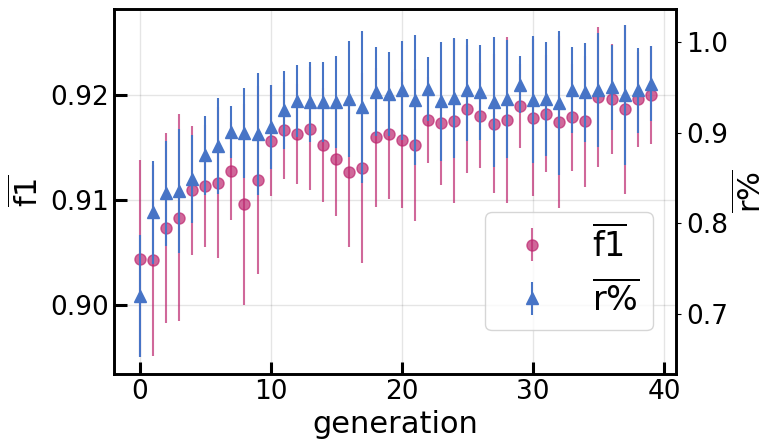}}
\caption{Average F1-score and proportion of max compression for English-EWT (a) and Japanese-GSD (b) during evolutionary search for optimal chunk type candidates.}
\label{fig:conv}
\end{figure}

We used a small network \carlos{for chunking} 
\carlos{due to the considerable computational costs of evolutionary algorithms.}
For each individual in each population, we trained a chunker for 5 epochs 
\carlos{(see Table \ref{tab:chunker_hyperparameters} in  Appendix \ref{appendix:hyper} for the parameters)}
and the corresponding model's best performance on the development set was taken as that individual's fitness along with the proportion of the maximum compression rate, $r\%$: the proportion of the maximum rate was used to prevent the algorithm from generating rulesets 
that generated few chunks and therefore minimising the potential impact. The convergence over 40 generations for English-EWT and Japanese-GSD can be seen in Figure \ref{fig:conv}.

As a final step, we took the top 100 best rulesets from across generations and extracted the rules that appeared in at least 75\% and 95\% of these sets, as the evolutionary algorithm only managed to find a single set with a fairly low performance. \markda{Rulesets were obtained this way for each treebank, except} 
the rulesets extracted from English-EWT were subsequently used on the other English UD treebanks. The statistics for the resulting chunks for the respective test data can be seen in Table \ref{tab:chunk_details}.

\begin{table}
    \centering
    \small
    \begin{tabular}{l cc|cc}
          & \multicolumn{2}{c}{\textbf{\# rules}} & \multicolumn{2}{c}{\textbf{C/sent}}  \\
          & 75\% & 95\% & 75\% & 95\% \\
         en-ewt & 230 & 134 & 3.11 & 2.71 \\
         en-gum & - & - & 4.48 & 3.84 \\
         en-lines & - & - & 4.47 & 4.18 \\
         en-partut & - & - & 6.32 & 5.84 \\
         bg & 152 & 108 & 3.94 & 3.65 \\
         de & 135 & 106 & 4.05 & 3.90 \\
         ja & 184 & 130 & 6.83 & 6.70 \\
    \end{tabular}
    \caption{Chunking statistics on test data for each treebank used where \# rules is the number of rules in a ruleset for a given threshold and C/sent corresponds to the number of chunks per sentence found.}
    \label{tab:chunk_details}
\end{table}

\section{Sequence-labelling framework}

All the proposed tasks can be cast as sequence labelling, 
\carlos{so}
in this work we \markda{have used} 
a sequence-labelling framework to address them. In particular, we \markda{rely}  
on bidirectional long short-term memory (BiLSTMs) networks \cite{hochreiter1997long,schuster1997bidirectional}. The input to the network are continuous word representations and character embeddings.

In this paper we use\markda{d} NCRF++ \cite{yang2018ncrf}, which uses stacked BiLSTMs, to generate contextuali\markda{s}ed hidden representations for every word ($\vec{h_i}$) in the input sentence. For decoding, it uses a feed-forward layer followed by a $\mathit{softmax}$ activation\markda{:} 
\begin{equation}
    \markda{P(y|\vec{h_i}) = \mathit{softmax}(\vec{W} \times \vec{h_i} + \vec{b})}
\end{equation}
\markda{The single task models are optimised with} 
cross-entropy loss, \markda{$\mathcal{L}$, defined as:}
\begin{equation}
    \markda{\mathcal{L}= -\sum log(P(y|h_i)}
\end{equation}
For the multi-task learning models, we implemented a hard-sharing architecture, where all the stacked BiLSTMs are shared across all tasks \cite{sogaard-goldberg-2016-deep}
. A separate feed-forward layer (as the one used in the single task setup) is used to decode the output for each task. With respect to the computation of the loss under the  \markda{multi-task learning} (MTL) setup, \markda{$\mathcal{L}_{MTL}$, is defined as}:
\begin{equation}
\markda{\mathcal{L}_{MTL}= \sum_{t \in T} \beta_t\mathcal{L}_t}
\end{equation}
\markda{where $t$ is a task from the set of all tasks, $T$; $\beta _t$ is the corresponding weight for task $t$; and $\mathcal{L}_t$ is the cross-entropy loss for task $t$.} A schematic of the network can be seen in Figure \ref{fig:mlt_net}.
\subsection{Dependency parsing as sequence labelling}\label{seqdp}
\begin{figure}
    \centering
    \includegraphics[width=0.99\linewidth]{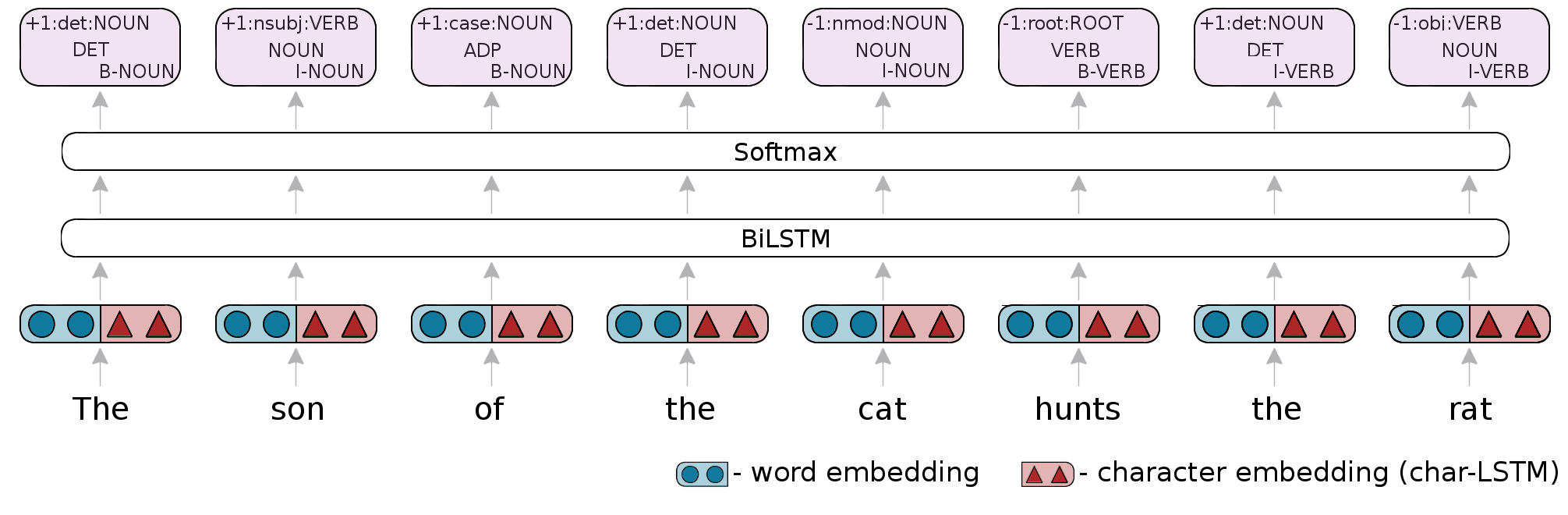}
    \caption{Multi-task architecture shown with sequence-labelling dependency parsing (as described in subsection \ref{seqdp}), POS tagging, and chunking as shared tasks. Network input is a concatenation of word embeddings (circles) and character-level word embeddings (triangles) obtained from a character-based LSTM layer. The network is constructed of BiLSTM layers followed by a $\mathit{softmax}$ layer for inference.}
    \label{fig:mlt_net}
\end{figure}

In order to \markda{more readily} utilise the multi-task framework for dependency parsing, we have cast dependency parsing as a sequence-labelling task. This was done by using the relative position encoding scheme introduced by \newcite{strzyz2019viable}. We opted to use this encoding as it was the highest performing labelling scheme they evaluated. For each word in a sentence the dependency relation label is combined with the relative position of its head based on the POS tag of the head, e.g. a noun which is the subject of a verb (\emph{son} in the input sentence in Figure \ref{fig:mlt_net}) would have a label of +1,nsubj,VERB, where +1 indicates the head is the next VERB in the sentence and nsubj is the relation label. 

\section{Experiments}
\paragraph{Data} The analyses were undertaken using the English treebanks (EWT, GUM, LinES, and ParTUT) and also Bulgarian-BTB, German-GSD, and Japanese-GSD from UD v2.3 \cite{ud23}. No results are given for Japanese-GSD for morphological feature tagging as it does not contain this information.
\paragraph{Network hyperparameters}
We used the framework \markda{as described above} and hyperparameters from \newcite{vilares2019better} 
which can be seen 
in Table \ref{tab:experimental_hyperparameters} \carlos{in the Appendix \ref{appendix:hyper}}. The standard input to the system consisted of word embeddings concatenated with character embeddings. 
All embeddings were randomly initialised.
\paragraph{Experiment 1}
We tested the impact of our chunks on POS and morphological feature tagging in a shared multi-task setting. This entails feeding word and character embeddings as input to the network with the output being some combination of POS tags, morphological feature tags, and chunk labels. These results were compared against the baseline taggers (single-task networks and POS and morphological features shared only). Tasks were equally weighted. As a further baseline we include results for POS and morphological feature tagging using UDPipe 2.2 \cite{straka2019}.
\paragraph{Experiment 2}
We used the best predictions (when using chunking) from experiment 1 as additional features for a sequence-labelling dependency parser \carlos{\cite{strzyz2019viable}}. Therefore, network input consisted of word and character embedding and then some combination of POS tags, morphological feature tags, or chunk labels with the sole output being a dependency parser tag. We used gold tags and labels as input during training, but at runtime we used predicted tags and labels. For baselines we train a model with no features which is decoded with predicted POS tags using UDPipe 2.2 (as the sequence-labelling encoding we are using requires POS tags to resolve dependency heads) and also a model trained with POS tags as features but also using UDPipe 2.2 predicted POS tags at runtime.
\paragraph{Experiment 3}
We tested the impact of our chunks on a sequence-labelling dependency parser in a multi-task framework with and without the other tasks. POS tagging was treated as a secondary main task with a weight of 0.5 (as POS tags are needed to decode the sequence-labelling scheme for the dependency parser) and chunks and morphological features were considered auxiliary tasks with a weight of 0.25 when used. The input during this experiment were only word and character embeddings. An example is shown in Figure \ref{fig:mlt_net} where the shared tasks are chunking, POS tagging, and dependency parsing. The baseline used here is a model trained solely to predict dependency parsing tags which are then decoded using predicted POS tags from UDPipe 2.2.

\section{Results and discussion}
\begin{table}
\small
    \centering
    \begin{tabular}{l cc|cc|cc|cc}
    \multicolumn{1}{c}{}&  \multicolumn{2}{c}{\textbf{ewt}} & \multicolumn{2}{c}{\textbf{gum}} & \multicolumn{2}{c}{\textbf{lines}} & \multicolumn{2}{c}{\textbf{partut}} \\
    & pos & feats & pos & feats & pos & feats & pos & feats \\
    udpipe & 94.44 & 95.37 & 93.88 & 94.21 & 94.73 & 94.83 & 94.10 & 94.01 \\ 
    single & 95.08 & 96.09 & 94.61 & 94.92 & 95.64 & 95.57 & 94.69 & 94.54 \\ 
    pos+feats & 95.23 & 96.21 & 94.60 & 95.26 & 95.59 & 95.71 & 94.63 & 94.16  \\\hline
      pos+feats+chunks$_{75}$   & \textbf{95.89} & \textbf{96.72} & \textbf{95.58} & \textbf{96.31} & \textbf{96.38} & \textbf{96.45} & 96.04 & 95.51 \\
        pos+feats+chunks$_{95}$   & 95.86 & 96.52 & 95.52 & 96.21 & 96.35 & 96.33 & \textbf{96.21} & \textbf{95.60} \\
    \end{tabular}
    \\\vspace{1em}
    \begin{tabular}{l cc|cc|cc}
    \multicolumn{1}{c}{}&  \multicolumn{2}{c}{\textbf{bg}} & \multicolumn{2}{c}{\textbf{de}} & \multicolumn{2}{c}{\textbf{ja}}  \\
    & pos & feats & pos & feats & pos & feats  \\
    udpipe & \textbf{97.78} & \textbf{95.55} & 92.03 & 70.18 & 96.39 & -  \\
    single & 97.41 & 95.06 & 93.07 & 87.14 & 96.97 & -  \\ 
    pos+feats & 97.69 & 94.84 & 92.90 & \textbf{87.28} & - & - \\\hline
    pos+feats+chunks$_{75}$ & 97.49 & 94.58 & \textbf{93.34} & 87.03 & 96.98 & -  \\
    pos+feats+chunks$_{95}$ & 97.44 &  94.45 & 92.90 & 87.11 & \textbf{97.09} & -  \\
    \end{tabular}
    \caption{
    Multi-task tagging performance on English UD treebanks (en-ewt, en-gum, en-lines, and en-partut), Bulgarian-BTB (bg), German-GSD (de), and Japanese-GSD (ja) UD treebanks: single, single-task training; pos, 
    with POS tagging; feats, 
    with morphological feature tagging (except Japanese (ja) which has no morphological features); and chunks$_x$, 
    with chunks with threshold $x$.}
    \label{tab:scores_tags}
\end{table}

\begin{table}[]
    \centering
    \small
    \begin{tabular}{l cc|cc}
    & \multicolumn{2}{c}{\textbf{baseline}} & \multicolumn{2}{c}{\textbf{multi}}\\
    & 75\% & 95\% & 75\% & 95\% \\
    en-ewt     & 89.99 & 91.59 & 91.84 & 92.98  \\
    en-gum     & 85.76 & 88.11 & 88.08 & 89.98  \\
    en-lines   & 86.01 & 88.38 & 88.45 & 90.67  \\
    en-partut  & 88.36 & 90.78 & 91.79 & 93.30  \\\hline
    bg      & 92.27 & 92.60 & 93.79 & 94.45  \\
    de      & 88.74 & 88.97 & 89.35 & 89.62  \\
    ja      & 93.35 & 92.73 & 94.39 & 94.02  \\
    \end{tabular}
    \caption{Chunker F1 scores in multi task setting where the baseline presented is from training the chunker for a given ruleset with threshold 75\% or 95\% as a single task and multi is from training with pos and morphological feature tagging except for Japanese (ja) which has no morphological features.}
    \label{tab:chunk_perform}
\end{table}
\begin{table}
    \centering
    \small
    \begin{tabular}{lcc|cc|cc|cc}
        \multicolumn{1}{c}{} &  \multicolumn{2}{c}{\textbf{en-ewt}} &  \multicolumn{2}{c}{\textbf{en-gum}} &  \multicolumn{2}{c}{\textbf{en-lines}} &  \multicolumn{2}{c}{\textbf{en-partut}} \\
           \multicolumn{1}{c}{}  & \multicolumn{1}{c}{uas} & \multicolumn{1}{c}{las} & \multicolumn{1}{c}{uas} & \multicolumn{1}{c}{las} & \multicolumn{1}{c}{uas} & \multicolumn{1}{c}{las} & \multicolumn{1}{c}{uas} & \multicolumn{1}{c}{las} \\
        no features$^{udpipe}$ & 80.97 & 77.87 & 76.70 & 72.71 & 76.43 & 71.87 & 81.63 & 78.67 \\ 
        pos$^{udpipe}$ & 84.88 & 81.79 & 81.09 & 76.87 & 79.06 & 74.08 & 84.01 & 80.63 \\ 
        \hline 
        pos & 86.15 & 83.29 & \textbf{83.03} & \textbf{79.31} & 80.76 & 76.12 & 85.83 & 82.69\\
        pos-feats & 86.32 & 83.37 & 82.83 & 79.13 & \textbf{81.15} & \textbf{76.48} & 86.71 & 83.60\\
        \hline
        pos-chunks$_{75}$ & 85.84 & 82.87 & 82.49 & 78.83 & 80.86 & 76.04 & 87.03 & 83.86\\
        pos-chunks$_{95}$ & 85.80 & 82.86 & 81.95 & 78.19 & 80.32 & 75.55 & 86.65 & 83.36\\
        pos-feats-chunks$_{75}$ & \textbf{86.43} & \textbf{83.41} & 82.61 & 78.86 & 81.13 & 76.21 & 87.09 & 83.86\\
        pos-feats-chunks$_{95}$ & 85.99 & 83.04 & 82.15 & 78.50 & 80.82 & 76.09 & \textbf{87.35} & \textbf{84.04}\\
    \end{tabular}
    \\\vspace{1.em}
    \begin{tabular}{l cc|cc|cc}
        \multicolumn{1}{c}{} &  \multicolumn{2}{c}{\textbf{bg}} &  \multicolumn{2}{c}{\textbf{de}} &  \multicolumn{2}{c}{\textbf{ja}} \\
           \multicolumn{1}{c}{}  & \multicolumn{1}{c}{uas} & \multicolumn{1}{c}{las} & \multicolumn{1}{c}{uas} & \multicolumn{1}{c}{las} & \multicolumn{1}{c}{uas} & \multicolumn{1}{c}{las} \\
        no features$^{udpipe}$ & 86.49 & 82.43 & 63.20 & 58.86 & 89.96 & 88.43 \\
        pos$^{udpipe}$ & 89.48 & 85.30 & 79.39 & 74.04 & 92.49 & 90.42 \\
        \hline 
        pos & 89.47 & 85.11 & 81.77 & 76.69 & \textbf{93.68} & \textbf{91.70}\\    
        pos-feats & \textbf{89.74} & \textbf{85.48} & \textbf{82.05} & \textbf{77.12} & - &-\\
        \hline
        pos-chunks$_{75}$ & 89.23 & 84.67 & 81.49 & 76.54 & 93.28 & 91.41\\
        pos-chunks$_{95}$ & 89.06 & 84.77 & 81.55 & 76.40 & 92.95 & 91.20\\
        pos-feats-chunks$_{75}$ & 89.11 & 84.83 & 81.77 & 76.71 & - &-\\
        pos-feats-chunks$_{95}$ & 89.24 & 85.07 & 81.41 & 76.38 & - &-\\
    \end{tabular}
    \caption{Feature input ablation for dependency parser with English UD treebanks (en-ewt, en-gum, en-lines, and en-partut), Bulgarian-BTB (bg), German-GSD (de), and Japanese-GSD (ja) UD treebanks: no features$^{udpipe}$, 
    no features but UDPipe predicted POS tags used to decode
    ; pos, gold POS tags 
    for training and predicted POS tags 
    for runtime (pos$^{udpipe}$ UDPipe predicted POS tags used); feats, gold morphological feature tags 
    for training and predicted feature tags 
    for runtime; and chunks$_x$, gold chunks with threshold $x$ 
    at training time and predicted chunks 
    for runtime.}
    \label{tab:full_ablation_en}
\end{table}
\begin{table}
    \centering
    \small
   \begin{tabular}{l cc|cc|cc|cc}
        \multicolumn{1}{c}{} &  \multicolumn{2}{c}{\textbf{en-ewt}} &  \multicolumn{2}{c}{\textbf{en-gum}} &  \multicolumn{2}{c}{\textbf{en-lines}} &  \multicolumn{2}{c}{\textbf{en-partut}} \\
           \multicolumn{1}{c}{}  & \multicolumn{1}{c}{uas} & \multicolumn{1}{c}{las} & \multicolumn{1}{c}{uas} & \multicolumn{1}{c}{las} & \multicolumn{1}{c}{uas} & \multicolumn{1}{c}{las} & \multicolumn{1}{c}{uas} & \multicolumn{1}{c}{las} \\
   single$^{udpipe}$ & 80.97 & 77.87 & 76.70 & 72.71 & 76.43 & 71.87 & 81.63 & 78.67 \\ 
   pos & 84.52 & 81.30 & 78.94 & 74.96 & 78.75 & 74.13 & 83.66 & 80.25\\
   pos-feats & 84.21 & 81.14 & 79.51 & 75.42 & 78.56 & 73.87 & 84.10 & 81.31\\\hline
   pos-chunks$_{75}$ & \textbf{84.55} & \textbf{81.51} & 79.54 & 75.48 & 78.17 & 73.55 & 83.86 & 81.13\\
   pos-chunks$_{95}$ & 84.42 & 81.34 & 79.60 & 75.54 & 78.72 & \textbf{74.20} & 83.57 & 80.16\\
   pos-feats-chunks$_{75}$ & 84.25 & 81.24 & \textbf{79.81} & \textbf{75.84} & 78.75 & 73.95 & 84.01 & 80.90\\
   pos-feats-chunks$_{95}$ & 84.24 & 81.18 & 79.48 & 75.36 & \textbf{78.84} & 74.15 & \textbf{84.98} & \textbf{81.92}\\
     \end{tabular}
     \\\vspace{1em}
 \begin{tabular}{l cc|cc|cc}
        \multicolumn{1}{c}{} &  \multicolumn{2}{c}{\textbf{bg}} &  \multicolumn{2}{c}{\textbf{de}}  &  \multicolumn{2}{c}{\textbf{ja}} \\
           \multicolumn{1}{c}{}  & \multicolumn{1}{c}{uas} & \multicolumn{1}{c}{las} & \multicolumn{1}{c}{uas} & \multicolumn{1}{c}{las} & \multicolumn{1}{c}{uas} & \multicolumn{1}{c}{las}\\
    single$^{udpipe}$ & 86.49 & 82.43 & 63.20 & 58.86 & 89.96 & 88.43 \\ 
   pos & 88.00 & 83.89 & 80.75 & 75.59 & \textbf{93.25} & 91.45\\
   pos-feats & 88.07 & 83.89 & 80.46 & 75.50 & - & -\\\hline
   pos-chunks$_{75}$ & 87.90 & 83.66 & \textbf{81.29} & \textbf{75.96} & \textbf{93.25} & \textbf{91.61}\\
   pos-chunks-$_{95}$ & 88.07 & 83.93 & 80.98 & 75.71 & 93.04 & 91.28 \\
   pos-feats-chunks$_{75}$ & \textbf{88.26} & \textbf{84.00} & 80.77 & 75.52 & - & -\\
   pos-feats-chunks$_{95}$ & 88.09 & 83.67 & 80.69 & 75.63 & - & -\\
  
    \end{tabular}
    \caption{Multi-task parsing results for English (en-ewt, en-gum, en-lines, and en-partut), Bulgarian-BTB (bg), German-GSD (de), and Japanese-GSD (ja) UD treebanks: single$^{udpipe}$, parsing as single task with UDPipe predicted POS tags used to decode 
    parser output; pos, 
    with POS tagging as aux. task; feats, 
    with morphological feature tagging as aux. task; and chunks$_x$, 
    with chunking as aux. task for threshold $x$.}
    \label{tab:multi_en}
\end{table}

As seen in Table \ref{tab:scores_tags} the multi-task framework with chunks improves the performance of both POS and morphological tagging for all English treebanks. In the same table, it is clear that they do not aid Bulgarian, but they do improve POS tagging performance for German and Japanese. 
Table \ref{tab:chunk_perform} shows that chunking performance consistently improves in the multi-task setting. 
\begin{figure}[htbp!]
    \centering
    \includegraphics[width=0.9\linewidth]{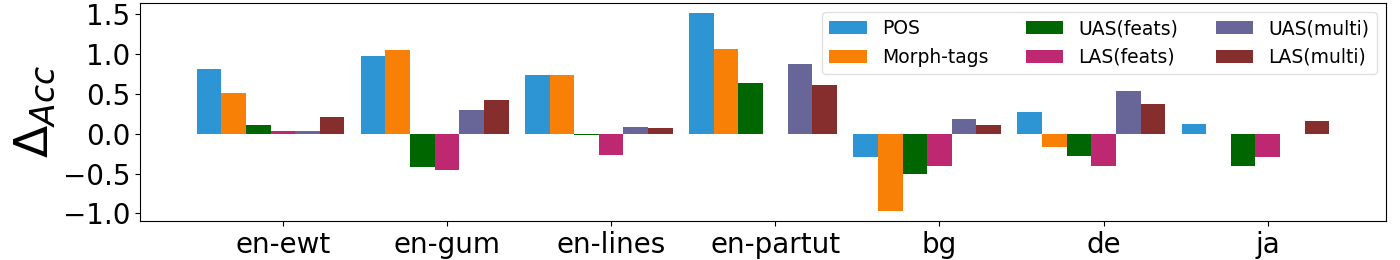}
    \caption{Difference in accuracy for each task between the best model with chunks and the best without.}
    \label{fig:delta_accs}
\end{figure}
Parsing
performance is improved across all treebanks when the predictions from experiment 1 are used as features (Table \ref{tab:full_ablation_en}), but only for English-EWT (the largest treebank) and ParTUT (the smallest) do the predicted chunks explicitly improve performance and for the other 
treebanks only the other predicted features help. This is in contrast to the findings of \newcite{nguyen2018improved}, who obtained higher performance for larger treebanks. 
In the multi-task setting for the dependency parser (Table \ref{tab:multi_en}), the chunking information consistently aids \markda{performance with a meaningful increase in accuracy observed over baseline models for each treebank.}

As can be seen in Figure \ref{fig:delta_accs}, the change in performance when using the predicted chunks as a feature for parsing is less profound than 
\carlos{in}
the multi-task experiments. Only two English treebanks explicitly benefit from predicted chunks, whereas all treebanks benefit from at least one feature. So the performance is at least implicitly improved by using our chunks, 
except for the more morphologically-rich (especially with respect to verbal inflection) Bulgarian. The treebank used for Japanese, generally an agglutinative language, does not contain morphological features, so perhaps it too would not improve with chunks if they could have been used. 
\carlos{Therefore,} 
it would be interesting to evaluate whether the impact of chunking information is predicated by certain linguistic features. Furthermore, 
the increase in performance for each treebank 
for the multi-task experiments suggests that the performance when using the chunks as input would improve 
with better predicted chunks, 
which corroborates the findings of \newcite{lacroix2018investigating}.


\section{Conclusion}
We have introduced a language-agnostic method for extracting chunks from dependency treebanks. We have also shown the efficacy of these chunks with respect to improving POS tagging, morphological feature tagging, and dependency parsing for a number of 
UD treebanks. 

\section*{Acknowledgments}

This work has received funding from the European Research Council (ERC), under the European
Union’s Horizon 2020 research and innovation programme (FASTPARSE, grant agreement No 714150), from the 
ANSWER-ASAP project (TIN2017-85160-C2-1-R) from MINECO, and from Xunta de Galicia (ED431B 2017/01). We thank one anonymous reviewer for in-depth comments and suggestions.

\clearpage
\bibliography{emnlp2018}
\bibliographystyle{acl}
\appendices
\clearpage
\section{Evolutionary algorithm}\label{appendix:algo}
\begin{algorithm} \caption{Evolutionary algorithm}\label{evo_algo}
\begin{algorithmic}[1]
\For{gen $\leftarrow$ max$_{\textrm{gen}}$}
\For{ind \textbf{in} population}
    \State ind.fit $\leftarrow$ \Call{GetFitness}{ind}
\EndFor
\State offspring $\leftarrow$ \Call{select}{population}
\State offspring $\leftarrow$ \Call{clone}{offspring}
\For{pair \textbf{in} offspring$_{\textrm{2i}}$, offspring$_{\textrm{2i+1}}$ }
    \If{random $<$ P$_\textrm{crossover}$}
    \State pair $\leftarrow$ \Call{Crossover}{pair}
    \EndIf
\EndFor
\For{ind \textbf{in} offspring}
    \If{random $<$ P$_\textrm{mutate}$}
    \State ind $\leftarrow$ \Call{Mutate}{ind}
    \EndIf
\EndFor
\State population $\leftarrow$ offspring
\EndFor
\Statex
\Function{GetFitness}{ind} 
    \State rules $\leftarrow$ \Call{Convert}{ind} 
    \State train, dev $\leftarrow$ \Call{ChunkTreebanks}{rules}
    \State \Call{TrainChunker}{train}
    \State F1 $\leftarrow$ \Call{EvalulateChunker}{dev}
    \State Rp $\leftarrow$ \Call{GetMaxRProportion}{dev}
    \State \Return F1 + 0.5$\cdot$Rp
\EndFunction
\end{algorithmic}
\end{algorithm}
\section{Hyperparameters}\label{appendix:hyper}
\begin{table}[H]
\centering
\tabcolsep=1cm
\small
\begin{tabular}{lr}
\textbf{hyperparameter} & \textbf{value}\\
     population size & 100  \\
     number of generations & 4\\
     k-best & 5\\
     P$_{\textrm{mutate}}$ & 0.5 \\
     P$_{\textrm{mutate gene}}$ & 0.05 \\
     P$_{\textrm{crossover}}$ & 0.5 \\
     decay (linear) & 0.1 \\
\end{tabular}
\caption{Hyperparameters for the evolutionary algorithm: k-best, the number of best parents chosen to seed next generation; P$_{\textrm{mutate}}$, the probability an individual will mutate; P$_{\textrm{mutate gene}}$, the probability a given gene will mutate;  P$_{\textrm{crossover}}$, the probability a pair of individuals will crossover; and decay is how much  P$_{\textrm{mutate}}$ and  P$_{\textrm{crossover}}$ decrease after each generation.}
\label{tab:evo_parameters}
\end{table}
\begin{table}[H]
\small
    \centering
        \tabcolsep=1.25cm  
    \begin{tabular}{l r}
    \textbf{hyperparameter} & \textbf{value}\\
         BiLSTM dimensions & 200  \\
         BiLSTM layers & 1 \\
         word embedding dimensions & 50\\
        character embedding dimensions & 30 \\
        character hidden dimensions &50 \\
        character CNN layers & 4 \\
        CNN window size & 3 \\
         optimiser & SGD \\
         loss function & cross entropy \\
         learning rate & 0.015 \\
        decay (linear) & 0.05 \\
        momentum & 0.9 \\
        dropout & 0.5 \\
        L$_2$ regularisation & 1$x10^{-8}$\\
        epochs & 5\\
        training batch size & 10\\
        runtime batch size & 128
    \end{tabular}
    \caption{Hyperparameters for the neural-net chunker used during the evolutionary algorithm.}
    \label{tab:chunker_hyperparameters}
\end{table}
\begin{table}[H]
\small
    \centering
        \tabcolsep=1.25cm  
    \begin{tabular}{l r}
    \textbf{hyperparameter} & \textbf{value}\\
         BiLSTM dimensions & 800  \\
         BiLSTM layers & 2 \\
         word embedding dimensions & 100\\
         character embedding dimensions & 30 \\
        character hidden dimensions &50 \\
        feature dimensions & 20\\
         optimiser & SGD \\
         loss function & cross entropy \\
         learning rate & 0.2 \\
        decay (linear) & 0.05 \\
        momentum & 0.9 \\
        dropout & 0.5 \\
        epochs & 100\\
        training batch size & 8\\
        runtime batch size & 128
    \end{tabular}
    \caption{Hyperparameters for the network used in all experiments.}
    \label{tab:experimental_hyperparameters}
\end{table}
\end{document}